\documentclass{iscram}

\usepackage[utf8]{inputenc}

\bibliography{main}

\usepackage{lipsum}
\usepackage{tikz}
\usepackage{fancyvrb}
\usepackage{listings}
\usepackage{enumitem}
\usepackage{tcolorbox}
\usepackage{multicol}
\usepackage{siunitx}
\usepackage{multirow}
\usepackage{graphicx}
\usepackage{booktabs}
\usepackage{balance}
\usepackage{subfig}
\usepackage[show]{inotes}
\usepackage{multirow, makecell}
\usepackage{pifont}%

\usepackage{hyperref}
\hypersetup{colorlinks,allcolors=black}

\newcommand{\spara}[1]{\smallskip\noindent{\bf #1}}
\newcommand{\cmark}{\ding{51}}%
\newcommand{\xmark}{\ding{55}}%

\lstdefinestyle{common}{
  xleftmargin=.5em,
  xrightmargin=.5em,
  frame=single,framesep=.5em,framerule=0pt,
  fancyvrb=true,
  basicstyle=\ttfamily,
  keywordstyle=\color{cyan!50!blue!75!black}\bfseries,
  commentstyle=\color{red!50!black}\itshape,
  stringstyle=\ttfamily\color{green!50!black},
  numbers=none,
  showspaces=false,
  showstringspaces=false,
  fontadjust=true,
  keepspaces=true,
  flexiblecolumns=true,
  emphstyle=\color{red},
}
\lstdefinestyle{TeX}{
  style=common,
  backgroundcolor=\color{blue!5},
  aboveskip=5pt,
  belowskip=5pt,
  language=[LaTeX]TeX,
  moretexcs={
    abstract, addbibresource, iscramset, keywords, mainmatter,
    maketitle, printbibliography, subsection, subsubsection, url,
    urldef, href, includegraphics, ldots, parencite, citeauthor,
    citeyear, citetitle, midrule, toprule, bottomrule
  },
  fancyvrb=true,
}
\lstdefinestyle{console}{
  style=common,
  backgroundcolor=\color{gray!10},
  aboveskip=5pt,
  belowskip=5pt,
}

\newlist{options}{description}{1}
\setlist[options]{%
  beginpenalty=10000,%
  itemsep=.5\parskip plus .3\parskip minus .2\parskip,
  parsep=.5\parskip plus .3\parskip minus .2\parskip,
  topsep=.5\parskip plus .3\parskip minus .2\parskip,
  partopsep=.5\parskip plus .3\parskip minus .2\parskip,
  style=nextline,labelindent=1em,%
  font=\normalfont\ttfamily}

\colorlet{macro color}{cyan!50!blue!75!black}
\colorlet{option color}{red!50!black}
\colorlet{generic color}{green!40!black}

\newtcolorbox{pseudoTeX}{colback=blue!5,colframe=blue!5,before=\nobreak}
\let\LaTeXorig\LaTeX
\renewcommand\LaTeX{\bgroup\fontfamily{lmr}\selectfont\upshape\LaTeXorig\egroup}

\urldef{\sitewebgind}\url{http://gind.mines-albi.fr}
\urldef{\sitewebminesalbi}\url{http://www.mines-albi.fr}
\urldef{\gmailpaulgaborit}\url{paul.gaborit@gmail.com}
\urldef{\jdmail}\url{j.doe@example.com}
\urldef{\sitex}\url{www.example.com}

\iscramset{
  CoRe Paper 2020={Social Media for Disaster Response and Resilience},
  title={Analysis of Social Media Data using\\Multimodal Deep Learning for\\Disaster Response},
  short title={Multimodal Deep Learning for Disaster Response},
  author={
    short name={Ofli},
    full name=Ferda Ofli,
    affiliation={
      Qatar Computing Research Institute\\
      Hamad Bin Khalifa University\\
      Doha, Qatar\\
      \href{mailto:fofli@hbku.edu.qa}{fofli@hbku.edu.qa}
    },
  },
  author={
    full name=Firoj Alam,
    affiliation={
      Qatar Computing Research Institute\\
      Hamad Bin Khalifa University\\
      Doha, Qatar\\
      \href{mailto:fialam@hbku.edu.qa}{fialam@hbku.edu.qa}
    },
  },
  author={
    full name=Muhammad Imran,
    affiliation={
      Qatar Computing Research Institute\\
      Hamad Bin Khalifa University\\
      Doha, Qatar\\
      \href{mailto:mimran@hbku.edu.qa}{mimran@hbku.edu.qa}
    },
  },
}

\usepackage{tikzpagenodes}
\usetikzlibrary{fit}

\begin{document}

\maketitle

% \makeatletter
% {\centering\large\iscram@version{}\\\iscram@date\par}
% \makeatother

\abstract{
Multimedia content in social media platforms provides significant information during disaster events. The types of information shared include reports of injured or deceased people, infrastructure damage, and missing or found people, among others. Although many studies have shown the usefulness of both text and image content for disaster response purposes, the research has been mostly focused on analyzing only the text modality in the past. In this paper, we propose to use both text and image modalities of social media data to learn a joint representation using state-of-the-art deep learning techniques. Specifically, we utilize convolutional neural networks to define a multimodal deep learning architecture with a modality-agnostic shared representation. Extensive experiments on real-world disaster datasets show that the proposed multimodal architecture yields better performance than models trained using a single modality (e.g., either text or image).
}

\keywords{Multimodal deep learning, Multimedia content, Natural disasters, Crisis Computing, Social media}

\section{Introduction}
\label{sec:introduction}

Information from different modalities often brings complementary signals about a concept, an object, an event, and the like. Learning from these different modalities leads to more robust inference compared to learning from a single modality. Multimodal learning is a well-researched area and has been applied in many fields including audio-visual analysis (e.g., videos)~\parencite{Poria2016,Pereira2016}, cross-modal study~\parencite{Nagrani2018}, and speech processing (e.g., audio and transcriptions)~\parencite{Chowdhury2019}. Despite the successes of multimodal learning in other areas, limited focus has been given to multimodal social media data analysis until recently~\parencite{gautam2019multimodal}. In particular, using social media data for social good requires time-critical analysis of the multimedia content (e.g., textual messages, images, videos) posted during a disaster situation to help humanitarian organizations in preparedness, mitigation, response, and recovery efforts. 

\begin{figure*}[htbp!]
	\renewcommand{\arraystretch}{0.5} % this reduces the vertical spacing between rows
	\linespread{0.5}\selectfont\centering
	\resizebox{.90\linewidth}{!}{%
% 		\begin{tabular}{p{0.1cm} p{0.28\textwidth} p{0.28\textwidth} p{0.28\textwidth}}
		\begin{tabular}{p{0.28\textwidth} p{0.28\textwidth} p{0.28\textwidth}}
			%\begin{tabular}{l c c c}
% 			&
			\textbf{Hurricane Maria}
			&
			\textbf{California Wildfires}
			&
			\textbf{Mexico Earthquake}
			\\
% 			\raisebox{4.5\normalbaselineskip}[0pt][0pt]{\rotatebox[origin=c]{90}{\textbf{Informative}}}
% 			&
			\includegraphics[width=0.28\textwidth]{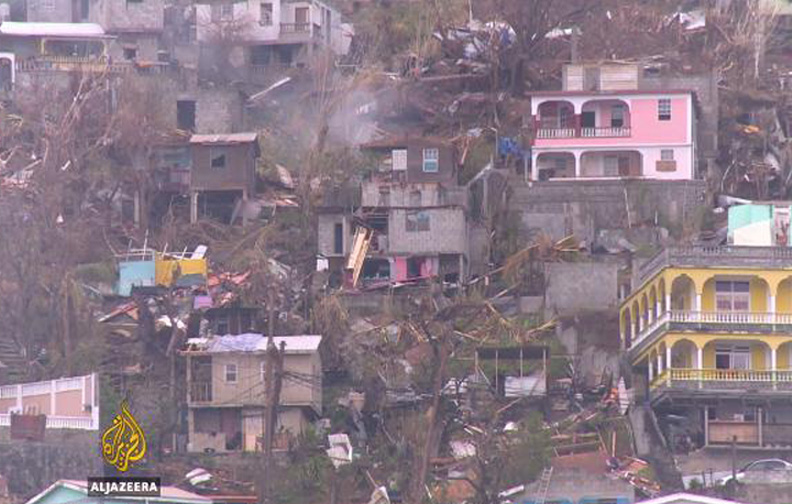}
			&
			\includegraphics[width=0.28\textwidth]{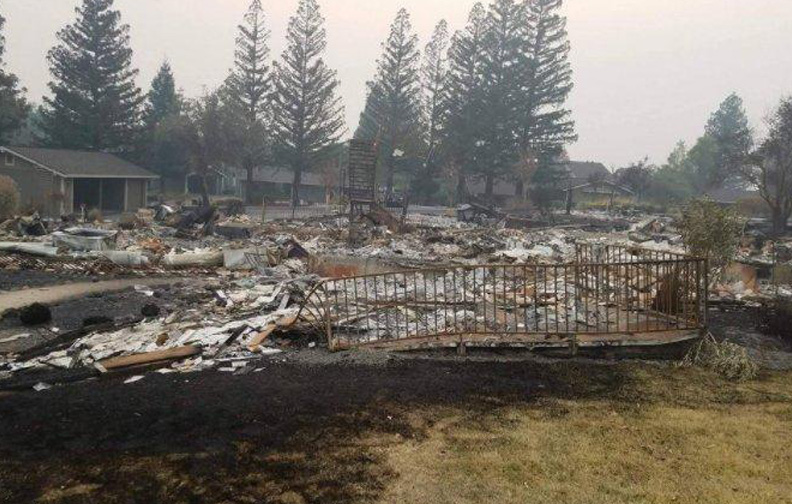}
			&
			\includegraphics[width=0.28\textwidth]{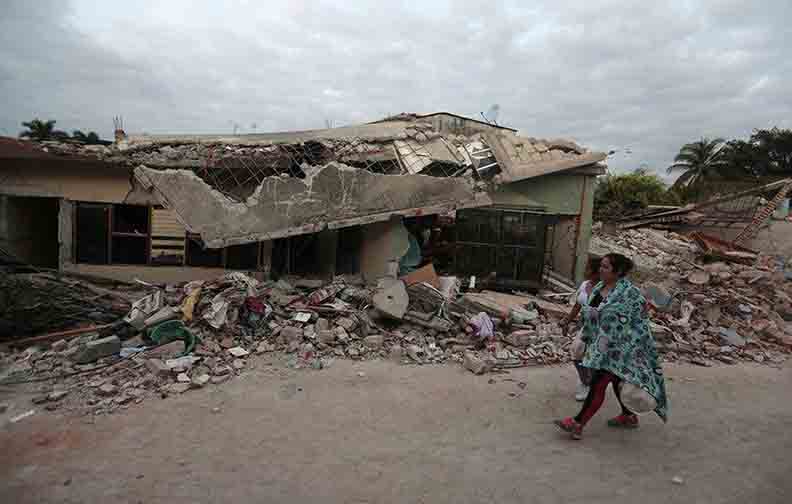}
			\\
% 			&
% 			{\scriptsize \textbf{(a)} Hurricane Maria turns Dominica into `giant debris field' https://t.co/rAISiAhMUy by \#AJEnglish via {@}c0nvey https://t.co/I4zeuW4gkc}
			{\scriptsize \textbf{(a)} Hurricane Maria turns Dominica into `giant debris field' https://t.co/rAISiAhMUy by \#AJEnglish via <USER>}
			&
% 			{\scriptsize \textbf{(b)} A friend's text message saved Sarasota man from deadly California wildfire https://t.co/0TNMFgL885 https://t.co/CIzo44Npza}
			{\scriptsize \textbf{(b)} A friend's text message saved Sarasota man from deadly California wildfire https://t.co/0TNMFgL885}
			&
% 			{\scriptsize \textbf{(c)} Earthquake leaves hundreds dead, crews combing through rubble in \#Mexico https://t.co/XPbAEIBcKw https://t.co/wGVxGD4xNd}
			{\scriptsize \textbf{(c)} Earthquake leaves hundreds dead, crews combing through rubble in \#Mexico https://t.co/XPbAEIBcKw}
			\\
% 			\raisebox{5.8\normalbaselineskip}[0pt][0pt]{\rotatebox[origin=c]{90}{\textbf{Not informative}}}
% 			&
% 			\includegraphics[width=0.28\textwidth]{images/maria_922757154557386752_0_not_informative.jpg}
% 			&
% 			\includegraphics[width=0.28\textwidth]{images/california_920896612859342849_0_not_informative.jpg}
% 			&
% 			\includegraphics[width=0.28\textwidth]{images/mexico_912123133968216065_0_not_informative.jpg}
% 			\\
% 			&
% 			{\scriptsize \textbf{(d)} @SueAikens hi su o back againe big hug FROM PUERTO RICO love you https://t.co/HCEyIHB0QZ}
% 			&
% 			{\scriptsize \textbf{(e)} https://t.co/jh0aQql3dR SEO ARTICLE GENERATOR https://t.co/2108RuhxgY \#blogging \#backlinks | Nurse fleeing California wildfires}
% 			&
% 			{\scriptsize \textbf{(f)} SEASON OVER???? WE COULD USE ABLE BODIES AT EARTHQUAKE IN MEXICO! DIG IN.... https://t.co/QlnYHtv9AI}
% 			\\
% 			\raisebox{5.2\normalbaselineskip}[0pt][0pt]{\rotatebox[origin=c]{90}{\textbf{Humanitarian}}}
% 			&
			\includegraphics[width=0.28\textwidth]{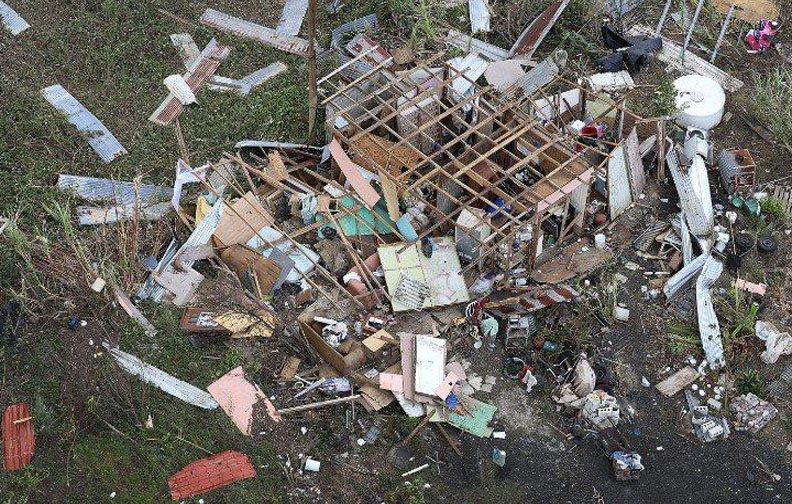}
			&
			\includegraphics[width=0.28\textwidth]{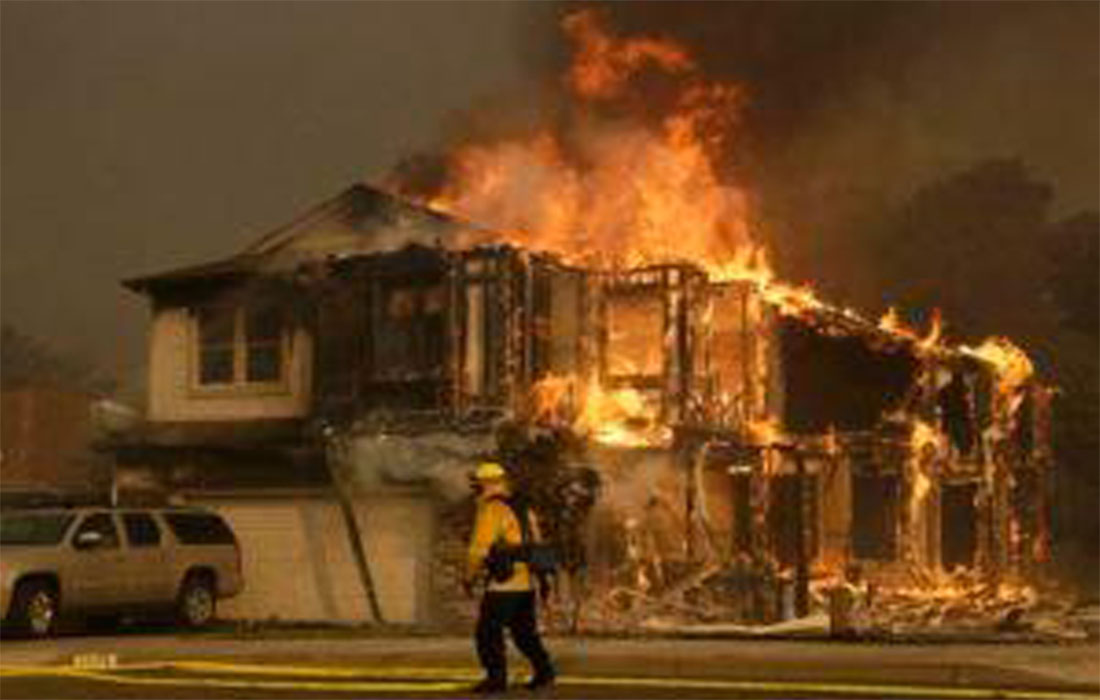}
			&
			\includegraphics[width=0.28\textwidth]{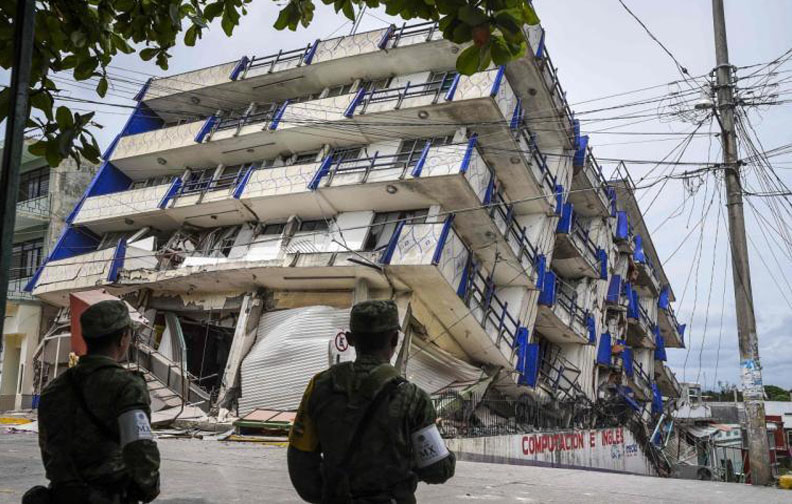}
			\\
% 			&
% 			{\scriptsize \textbf{(d)} Corporate donations for Hurricane Maria relief top \$24 million https://t.co/w34ZZziu88 https://t.co/ePddksfFc2}
			{\scriptsize \textbf{(d)} Corporate donations for Hurricane Maria relief top \$24 million https://t.co/w34ZZziu88}
			&
% 			{\scriptsize \textbf{(e)} California Wildfires Threaten Significant Losses for P/C Insurers, Moodya Says https://t.co/ELUaTkYbzZ https://t.co/Os8UAAjxGb}
			{\scriptsize \textbf{(e)} California Wildfires Threaten Significant Losses for P/C Insurers, Moodya Says https://t.co/ELUaTkYbzZ}
			&
% 			{\scriptsize \textbf{(f)} Southern Mexico rocked by 6.1-magnitude earthquake CLICK BELOW FOR FULL STORY... https://t.co/Vkz6fNVe5s... https://t.co/Cn4LSWrN4T}
			{\scriptsize \textbf{(f)} Southern Mexico rocked by 6.1-magnitude earthquake CLICK BELOW FOR FULL STORY... https://t.co/Vkz6fNVe5s...}
			\\
		\end{tabular}
	}
	\caption{Tweet text and image pairs from different disaster events with complementary information.}
	\label{fig:sample_crisis_images}
	%\vspace{-.2in}
\end{figure*}

Figure~\ref{fig:sample_crisis_images} shows a few tweets with associated images collected from three recent major disasters. We observe that relying on a single modality may often miss important insights. For instance, although the tweet text in Figure~\ref{fig:sample_crisis_images}(f) reports about a 6.1 magnitude earthquake in Southern Mexico, the scale of the damage incurred by this earthquake cannot be inferred from the text. However, if we analyze the image attached to this tweet, we can easily understand the immense destruction caused by the earthquake.

%Compared to these research areas, the study of multimodal social media analysis in crisis computing are relatively very few. Very recently its importance has been highlighted in some studies \parencite{alam2018crisismmd,Mouzannar2018}. In Figure \ref{fig:sample_crisis_images}, we provide a few examples from the work of \parencite{alam2018crisismmd} to highlight the fact that information from unimodality (i.e., either text or image in this use case) often miss important insights. For example, from the Figure \ref{fig:sample_crisis_images}(f), from the text we might not get the idea of damage that caused by the earthquake, however, image gives us clear evidence about the event.  

%Processing social media data to extract such information can significantly help humanitarian organizations in preparedness, response, and recovery of an emergency situation. This kind of information is useful for humanitarian organizations to deal with different challenges such as handling information overload, information classification and determining its credibility, and prioritizing certain types of information, etc.~\parencite{imran2015processing}. Hence it is important to develop automatic system that can deal with both textual and imaginary content and extract useful information to support decision makers in humanitarian organizations. 

Most of the previous studies that rely on social media for disaster response have mainly focused on textual content analysis, and little focus has been given to images shared on social media~\parencite{imran2015processing,CarlosCastillo2016}. Many past research works have demonstrated that images shared on social media during a disaster event can help humanitarian organizations in a number of ways. For example, \cite{nguyen17damage} use images shared on Twitter to assess the severity of infrastructure damage. \cite{Mouzannar2018} also focus on identifying damages in infrastructure and environmental elements. Taking a step further, \cite{gautam2019multimodal} have recently presented a work on multimodal analysis of crisis-related social media data for identifying informative tweet text and image pairs.

In this study, we also aim to use both text and image modalities of Twitter data to learn (i) whether a tweet is informative for humanitarian aid or not, and (ii) whether it contains some useful information such as a report of injured or deceased people, infrastructure damage, etc. We tackle this problem in two separate classification tasks and solve them using multimodal deep learning techniques.

%Humanitarian organizations face many challenges at the onset of a disaster situation. One important challenge is to gain situational awareness i.e., to determine the scale of damage, affected and injured people, urgent needs etc. We aim to use social media textual and imagery content to enhance this during a disaster
%One of the major limitations of multimodal study of social media data for crisis computing was a lack of labelled data. The study of \parencite{alam2018crisismmd,Mouzannar2018} has opened an window for the multimodal study in crisis computing. Data collection and annotation process is one of the major challenge for multimodal study due to the noisy nature of the social media data and the non-existence of both modalities. It also raises one interesting research question, how can we develop a multimodal architecture which can deal with input while one modality is absent?  

%% Typical approach to deal with multimodal study
The typical approach to deal with multimodality includes feature- or decision-level fusion, which is also termed as early and late fusion \parencite{kuncheva2004combining,alam2014}. In deep learning architectures, multimodality is combined at the hidden layers with a different variant of network architecture such as static, dynamic, and N-way classification as can be seen in \parencite{ngiam2011multimodal,Nagrani2018,Chowdhury2019}. Specifically, \cite{ngiam2011multimodal} explore different architectures for audio-visual data. Their study includes unimodal as well as cross-modal learning (i.e., learning one modality while giving multiple modalities during feature learning), multimodal fusion, and shared representation learning. \cite{Nagrani2018} also study audio-visual data for a biometric matching task while they investigate different deep neural network architectures for developing a multimodal network whereas \cite{Chowdhury2019} analyze audio and transcriptions while concatenating both modalities in a hidden layer. 

%% our contribution in this study
In this work, we propose to learn a joint representation from two parallel deep learning architectures where one architecture represents the text modality and the other architecture represents the image modality. For the image modality, we use the well-known VGG16 network architecture and extract high-level features of an input image using the penultimate fully-connected (i.e., fc2) layer of the network. For the text modality, we define a Convolutional Neural Network (CNN) with five hidden layers and different filters. Two feature vectors obtained from both modalities are then fed into a shared representation followed by a dense layer before performing a prediction using softmax. In the literature, this type of joint representation is also known as early fusion. The proposed multimodal architecture is trained using three different settings as follows: (i) train a network using input data from both modalities, (ii) train a network using only the text modality, and (iii) train a network using only the image modality.

% \todo[ferda]{below two paragraphs need to be updated according to the latest experimental setup.}

We perform extensive experiments on a real-world disaster-related dataset collected from Twitter, i.e,. CrisisMMD~\parencite{alam2018crisismmd}, using the aforementioned three training settings for two different classification tasks: informativeness and humanitarian classification. The test data for the evaluation of all three settings are fixed.
% with a variable number of training instances. 
% The experiments performed on a real-world disaster-related dataset collected from Twitter 
The experimental results show that the proposed approach (i.e., multimodal learning) outperforms our baseline models trained on a single modality (i.e., either text or image). For the informativeness classification task, our best model obtained an F1-score=84.2 and for the humanitarian classification task, our best model achieved an F1-score=78.3. Despite the fact that this model outperforms its counter-part unimodal baseline models (i.e., trained on a single modality), we remark that there is a big room for improvement, which we leave as future work. 
\textcolor{black}{To the best of our knowledge, this is the first study that presents baseline results on CrisisMMD using state-of-the-art deep learning-based unimodal and multimodal approaches for both informativeness and humanitarian tasks, all in one place.}
We hope that experimental analyses presented in this study will provide guidance for future research using the CrisisMMD dataset. %Thus, we remark that an extensive future work has to be conducted to further improve the performance of these models.

%In this study we used CrisisMMD dataset \parencite{alam2018crisismmd} in order to create baseline results for the research community. It is one of the largest multimodal social media dataset for disaster response. In terms of deep learning network architecture we used static architecture, in which hidden layers from both text and image network are combined into another layer followed by a few more hidden layer before softmax. In literature it is also referred to as early fusion. More details of the network are provided in experiment section. Towards the direction of multimodal study for disaster response out contributions of this study include, 1) preparing dataset by pairing the labels coming from both textual and imagenary content, 2) unimodal experiment,3) multimodal experiment with shared representation, 4) experiment with multimodal architecture while one modality is absent. Our study differs from previous studies in a way that we use one of the largest annotated datasets for classifying \textit{informativeness} and \textit{humanitarian categories} to extract actionable information. 

% The rest of the paper is organized as follows. In Section~\ref{sec:related_work}, we provide a summary of the related work. Then, in Section~\ref{sec:data_collection}, we provide details of the dataset we used in this study. Next, in Section~\ref{sec:experiments}, we discuss the experimental methodology and results. We then present possible applications and discussion in Section~\ref{sec:discussion}. Finally, we conclude the paper in Section~\ref{sec:conclusions}.

The rest of the paper is organized as follows. In the \textit{Related Work} section, we provide a review of the literature. Then, in the \textit{Dataset} section, we present details of the dataset used in this study. Next, in the \textit{Experiments} section, we describe the methodology and discuss experimental results. We then present possible applications and future directions in the \textit{Discussion} section. Finally, we conclude the paper with the \textit{Conclusions} section.

\section{Related Work}
\label{sec:related_work}

% \inote[ferda]{We may want to update this section to include anything recently published.} 

Many past studies have analyzed social media data, especially textual content, and demonstrated its useful for humanitarian aid purposes~\parencite{imran2015processing, CarlosCastillo2016}. With recent successes of deep learning, research works have started to use social media images for humanitarian aid. %In crisis computing text-based analyses of social media data are comparatively richer than imagery content analysis. Very recently the attention has been focused on imagery content and multimodal content analysis. 
For instance, the importance of imagery content on social media has been reported in many studies~\parencite{petersinvestigating,daly2016mining,chen2013understanding,nguyen2017automatic,nguyen17damage,alam17demo,alam2019SocialMedia}. \cite{petersinvestigating} analyzed the data collected from Flickr and Instagram for the flood event in Saxony, 2013. Their findings suggested that the existence of images within on-topic textual content were more relevant to the disaster event, and the imagery content also provided important information, which was related to the event. Similarly, \cite{daly2016mining} analyzed images extracted from social media data, which is focused on a fire event. They analyzed spatio-temporal meta-data associated with the images and suggested that geotagged information are useful to locate the fire affected areas. The analysis of imagery content shared on social media has been explored using deep learning techniques in several studies~\parencite{nguyen2017automatic,nguyen17damage,alam17demo}. Furthermore, \cite{alam2019SocialMedia} presented an image processing pipeline to extract meaningful information from social media images during a crisis situation, which has been developed using deep learning-based techniques. Their image processing pipeline includes collecting images, removing duplicates, filtering irrelevant images, and finally classifying them with damage severity.

Combining textual and visual content can provide highly relevant information as discussed by \cite{bica2017visual} where they explored social media images posted during two major earthquakes in Nepal during April-May 2015. Their study focused on identifying geo-tagged images and their associated damage. \cite{chen2013understanding} studied the association between tweets and images, and their use in classifying visually relevant and irrelevant tweets. They designed classifiers by combining features from the text, images and socially relevant contextual features (e.g., posting time, follower ratio, the number of comments, re-tweets), and reported an F1-score of $70.5\%$ in a binary classification task, which is $5.7\%$ higher than the text-only classification. Recently, \cite{Mouzannar2018} explored damage detection by focusing on human and environmental damages. Their study explores unimodal as well as different multimodal modeling setups based on a collection of multimodal social media posts labeled with six categories such as infrastructural damage (e.g., damaged buildings, wrecked cars, and destroyed bridges), damage to natural landscape (e.g., landslides, avalanches, and falling trees), fires (e.g., wildfires and building fires), floods (e.g., city, urban and rural), human injuries and deaths, and no damage.
% Their experimental methodology includes 1) unimodal, 2) multimodal with decision- and feature-level fusion, and shared representations. 
\textcolor{black}{Similarly, \cite{gautam2019multimodal} presented a comparison of unimodal and multimodal methods on crisis-related social media data using an approach based on decision fusion for classifying tweet text and image pairs into informative and non-informative categories.}

For the tweet classification task, deep learning-based techniques such as Convolutional Neural Networks (CNN)~\parencite{nguyen2017robust}, and Long-Short-Term-Memory Networks (LSTM)~\parencite{rosenthal2017semeval} have been widely used. For the image classification task, state-of-the-art works also utilize different techniques of deep neural networks such as Convolutional Neural Networks (CNN) with deep architectures. Among different CNN architectures, the most popular are VGG~\parencite{simonyan2014very}, AlexNet~\parencite{krizhevsky2012imagenet}, and GoogLeNet~\parencite{szegedy2015going}. The VGG is designed using an architecture with very small (3$\times$3) convolution filters and with a depth of 16 and 19 layers. The 16-layer network is referred to as VGG16 network, which we used in this study.

%% multimodal techniques/architecture
For combining multiple modalities, early and late fusion have been the traditional approaches \parencite{kuncheva2004combining}. The early-fusion approaches combine features from different modalities while the late-fusion approaches make classification decisions either by majority voting or stacking methods (i.e., combining classifiers' output and making a decision by training another classifier). In deep learning paradigm, typically, hidden layers are combined using approaches such as static or dynamic concatenation as discussed in \parencite{Nagrani2018,Chowdhury2019}. 
%\inote{we have to say how we are different or better than this work (as we cited them) “Damage Identification in Social Media Posts using Multimodal Deep Learning”}
In this study, we follow CNN-based deep learning architectures for both unimodal and multimodal experiments. We extract high-level features from two independent modality-specific networks (i.e., text and image) and concatenate them to form a shared representation for our classification tasks.

\textcolor{black}{Our study differs from previous studies in a number of ways~\parencite{Mouzannar2018,gautam2019multimodal}. As such, we experiment with one of the largest, publicly available datasets (i.e., CrisisMMD) to provide baseline results for two popular crisis response tasks, i.e., \textit{informativeness} and \textit{humanitarian categorization}, using multimodal deep learning with a feature-fusion approach. In contrast, \cite{Mouzannar2018} focus only on human and environmental damages using a home-grown dataset, which limits generalization of their findings. As for \cite{gautam2019multimodal}, although they also use a subset of the CrisisMMD dataset, they focus only on the informativeness classification task and employ a decision-fusion approach in their experiments. Unfortunately, due to potential differences in data organization (i.e., training, validation, and test splits), our experimental results are not exactly comparable with theirs.}

% In summary, our work complements the existing works in two ways: (i) we provide baseline results for the humanitarian categorization task in addition to the informativeness task, and (ii) our experimental results are based on an feature-fusion approach as opposed to a decision-fusion approach.

% \textcolor{red}{The study by \citeauthor{gautam2019multimodal} (\citeyear{gautam2019multimodal}) also uses CrisisMMD dataset and proposes a multimodal deep learning method to only classify informativeness of a tweet. Even though the dataset is the same, however, due to the different ways of data split the experimental results are not exactly comparable.}

\section{Dataset}
\label{sec:data_collection}

% \todo[ferda]{Update this section to explain why/how we use a subset where the image/text labels agree.}

We use CrisisMMD\footnote{\url{http://crisisnlp.qcri.org/}} dataset, which is a multimodal dataset consisting of tweets and associated images collected during seven different natural disasters that took place in 2017~\parencite{alam2018crisismmd}. The dataset is annotated for three tasks: (i) informative \textit{vs.} not-informative, (ii) humanitarian categories (eight classes), and (iii) damage severity (three classes). Because the third task, i.e., damage severity, was applied only on images, we do not consider this task in the current study and focus only on the first two tasks as follows. 
%  Annotations were run independently for each event using Figure Eight platform\footnote{\url{https://www.figure-eight.com/}} 

\spara{Informative vs.\ Not-informative.} The purpose of this task is to determine whether a given tweet text or image, collected during a disaster event, is useful for humanitarian aid purposes\footnote{A detailed definition of \emph{humanitarian aid} is provided in~\parencite{alam2018crisismmd}.}. If the given text (image) is useful for humanitarian aid, it is considered as an ``informative'' tweet (image), otherwise as a ``not-informative'' tweet (image). 
% The definition of \emph{humanitarian aid} is provided in~\parencite{alam2018crisismmd}.
% In this task, \textit{humanitarian aid} refers to any message (text or image) content that can provide assistance to people who need help. The primary purpose of \textit{humanitarian aid} is to save lives, reduce suffering, and rebuild affected communities. Affected people include homeless, refugees, and victims of natural disasters, wars, and conflicts with dire needs for basic necessities of life such as like food, water, shelter, medical assistance, and damage-free critical infrastructure and utilities (e.g., roads, bridges, power and communication lines). Moreover, the tweet/image is considered ``Informative'' if it reports/shows one or more of the following: caution, advice, or warning; injured, dead, or affected individuals; rescue, volunteering, or donation requests or efforts; damaged houses, damaged roads, or damaged buildings; flooded houses or flooded streets; blocked roads, blocked bridges, or blocked pathways; any built structure affected by earthquake, fire, heavy rain, strong winds, gust, etc. Content showing banners, logos, and cartoons are \emph{not} considered as ``Informative.''

\spara{Humanitarian Categories.} The purpose of this task is to understand the type of information shared in a tweet text/image, which was collected during a disaster event. Given a tweet text/image, the task is to categorize it into one of the categories listed in Table~\ref{table:data_distribution}.
% \begin{itemize}
% 	\item Infrastructure and utility damage
% 	\item Vehicle damage
% 	\item Rescue, volunteering, or donation effort
% 	\item Injured or dead people
% 	\item Affected individuals
% 	\item Missing or found people
% 	\item Other relevant information
% 	\item Not relevant or can't judge
% \end{itemize}

An important property of CrisisMMD is that some of the co-occurring tweet text and image pairs have different labels for the same task because text and image modalities were annotated separately and independently. Therefore, in this study, we consider only a subset of the original dataset where text and image pairs have the same label for a given task. As a result of this filtering, some of the categories in the humanitarian task are left with only a few pairs of tweet text and image. This situation skews the overall label distribution and creates a challenging setup for model training. To deal with this issue, we combine those minority categories that are semantically similar or relevant. Specifically, we merge the ``injured or dead people'' and ``missing or found people'' categories into the ``affected individuals'' category. Similarly, we merge ``vehicle damage'' category into the ``infrastructure and utility damage'' category. \textcolor{black}{As a result, we are left with five categories for the humanitarian task as shown in Table~\ref{table:data_distribution}.}

% \todo[ferda]{need to explain how we did the data split so that tweets with multiple images appear only in training set. our motivation for this was to have a consistent dev and test set between unimodal and multimodal experiments.}

%Another important property of the CrisisMMD dataset is that some tweets are associated with multiple images. Therefore, when we do the data split for training, development, and test sets, we need to make sure that the development and test sets are consistent across unimodal and multimodal experiments (i.e., there is no double counting of tweet texts). To achieve this, we simply assign all instances of tweet text with multiple images into the training set so that in the development and test sets we have the same number of \emph{unique} tweet texts, images, and their pairs. 

\textcolor{black}{Twitter allows attaching up to four images to a tweet, and hence, CrisisMMD contains some tweets that have more than one image, i.e., the same tweet text is associated with multiple images. While splitting data into training, development, and test sets, we need to make sure that no duplicate tweet text exists across these splits. To achieve this, we simply assign all tweets with multiple images only to the training set. This also ensures that there are no repeated data points (i.e., tweet text) in the development and test sets for the unimodal experiments on text modality.} While doing so, we maintain a 70\%:15\%:15\% data split ratio for training, development, and test sets, respectively. Table~\ref{table:data_distribution} provides the final set of categories, total number of tweet text and images in each category as well as their split into training, development, and test sets.\footnote{The data split used in the experiments can be found online at \url{http://crisisnlp.qcri.org/}.} Note that the total number of tweet text and images in the table differ only for the training sets as per the strategy explained above.

\begin{table*}
\centering
\caption{List of categories and their data split for different tasks.}
\label{table:data_distribution}
%\scalebox{0.8}{
\resizebox{\columnwidth}{!}{%
\begin{tabular}{@{}lrrrrrrrr@{}}
\toprule
 & \multicolumn{2}{c}{\textbf{Train (70\%)}} & \multicolumn{2}{c}{\textbf{Dev (15\%)}} & \multicolumn{2}{c}{\textbf{Test (15\%)}} & \multicolumn{2}{c}{\textbf{Total}} \\ \cmidrule(lr){2-3}\cmidrule(lr){4-5}\cmidrule(lr){6-7}\cmidrule(lr){8-9}
 & \multicolumn{1}{c}{\bf Text} & \multicolumn{1}{c}{\bf Image} & \multicolumn{1}{c}{\bf Text} & \multicolumn{1}{c}{\bf Image} & \multicolumn{1}{c}{\bf Text} & \multicolumn{1}{c}{\bf Image} & \multicolumn{1}{c}{\bf Text} & \multicolumn{1}{c}{\bf Image} \\ \midrule
\multicolumn{9}{@{}l}{\textbf{Informative Task}} \\
\textit{Informative} & 5,546 & 6,345 & 1,056 & 1,056 & 1,030 & 1,030 & 7,632 & 8,431 \\
\textit{Not-informative} & 2,747 & 3,256 & 517 & 517 & 504 & 504 & 3,768 & 4,277 \\ \midrule
\textit{Total} & 8,293 & 9,601 & 1,573 & 1,573 & 1,534 & 1,534 & 11,400 & 12,708 \\\midrule
\multicolumn{9}{@{}l}{\textbf{Humanitarian Task}} \\
\textit{Affected individuals} & 70 & 71 & 9 & 9 & 9 & 9 & 88 & 89 \\
\textit{Rescue volunteering or donation effort} & 762 & 912 & 149 & 149 & 126 & 126 & 1,037 & 1,187 \\
\textit{Infrastructure and utility damage} & 496 & 612 & 80 & 80 & 81 & 81 & 657 & 773 \\
\textit{Other relevant information} & 1,192 & 1,279 & 239 & 239 & 235 & 235 & 1,666 & 1,753 \\
\textit{Not-humanitarian} & 2,743 & 3,252 & 521 & 521 & 504 & 504 & 3,768 & 4,277 \\ \midrule
\textit{Total} & 5,263 & 6,126 & 998 & 998 & 955 & 955 & 7,216 & 8,079 \\ \bottomrule
\end{tabular}%
}
\end{table*}

\section{Experiments}
\label{sec:experiments}

% \begin{figure*}[ht]
% \centering
% \scalebox{0.5}{
% \includegraphics{figs/multimodal_architecture.png}
% }
% \caption{The multimodal architecture for the classification task using both text and image as input to the system.}
% \label{fig:multimodal}
% \end{figure*}

% Given the two annotation tasks explained in the previous section, we treat them as two separate classification tasks. For each classification task, we design our experimental framework in three different settings as follows:

% \begin{enumerate}
%     \item Unimodal learning using texts (baseline)
%     \item Unimodal learning using images (baseline)
%     \item Multimodal learning using both text and images
% \end{enumerate}

As explained in the previous section, we have two sets of annotations for two separate classification tasks, i.e., informativeness and humanitarian. For each one of these tasks, we perform three classification experiments where we train models using (i) only tweet text, (ii) only tweet image, and (iii) tweet text and image together.

In the following subsections, we first describe the data preprocessing steps and then describe the deep learning architecture used for each modality as well as their training details. To measure the performance of the trained models, we use several well-known metrics such as accuracy, precision, recall, and F1-score.

\subsection{Data Preprocessing}
The textual content of tweets is often noisy, usually consisting of many symbols, emoticons, and invisible characters. Therefore, we preprocess them by removing stop words, non-ASCII characters, numbers, URLs, and hashtag signs. We also replace all punctuation marks with white-spaces.

On the image side, we follow the typical preprocessing steps of scaling the pixels of an image between 0 and 1 and then normalizing each channel with respect to the ImageNet dataset \parencite{deng2009imagenet}.

\subsection{CNN: Text Modality}
\label{ssec:text_classification}
%To design the classification model 
For the text modality, we use Convolutional Neural Network (CNN) due to its better performance in crisis-related tweet classification tasks reported in~\parencite{nguyen2017robust}. Specifically, we create a CNN consisting of 5 hidden layers. To input the network, we zero-padded the tweets for an equal length and then converted them into a word-level matrix where each row represents a word in the tweet extracted using a pre-trained word2vec model discussed in \parencite{Alam2018GraphBS}. This word2vec model is trained using the Continuous Bag-of-Words (CBOW) approach of \cite{mikolov2013efficient} on a large disaster-related dataset of size $364$ million tweets with vector dimensions of $300$, a context window size of 5 and $k=5$ negative samples.

The input then goes through a series of sequential layers including the convolutional layer, followed by the max-pooling layer, to obtain a higher-level fixed-size feature representation for each tweet. These fixed-size feature vectors are then passed through one or more fully connected hidden layers, followed by an output layer. In the convolutional and fully-connected layers, we use rectified linear units (ReLU) \parencite{krizhevsky2012imagenet} as the activation function, and in the output layer, we use the softmax activation function. 

We train the CNN models using the Adam optimizer~\parencite{zeiler2012adadelta}. The learning rate is set to $0.01$ when optimizing for the classification loss on the development set. The maximum number of epochs is set to 50, and dropout~\parencite{srivastava2014dropout} rate of $0.02$ is used to avoid overfitting. We set \emph{early-stopping} criterion based on the accuracy on the development set with the patience of 10. We use 100, 150, and 200 filters with the corresponding window size of 2, 3, and 4, respectively. We use the same pooling length as the filter window size. 
%We do not tune any hyperparameter (e.g., the size of hidden layers, filter size, dropout rate) in the experimental setting. 
We also apply batch normalization due to its success reported in the literature~\parencite{ioffe2015batch}.

\subsection{VGG16: Image Modality}
\label{ssec:image_classification}
% To design the image classification model
For the image modality, we employ a transfer learning approach, which is an effective approach for visual recognition tasks~\parencite{yosinski2014transferable,ozbulak2016transferable}. The idea of the transfer learning approach is to use existing weights of a pre-trained model. We use the weights of a VGG16 model pre-trained on ImageNet to initialize our model. We adapt the last layer (i.e., softmax layer) of the network according to the particular classification task at hand instead of the original 1,000-way classification. The transfer learning approach allows us to transfer the features and the parameters of the network from the broad domain (i.e., large-scale image classification) to the specific one, in our case informativeness and humanitarian classification tasks.

We train the image models using the Adam optimizer~\parencite{zeiler2012adadelta} with an initial learning rate of $10^{-6}$, which is reduced by a factor of $0.1$ when accuracy on the development set stops improving for 100 epochs. We set the maximum number of epochs to 1,000 with an early-stopping criterion. 

% \todo[ferda]{add more details about training configuration, e.g., optimizer, learning rate, batch size, batch norm, number of epochs, etc.}

\subsection{Multimodal: Text and Image}
\label{ssec:text_image_classification}
\begin{figure*}[]
\centering
\scalebox{0.6}{
\includegraphics{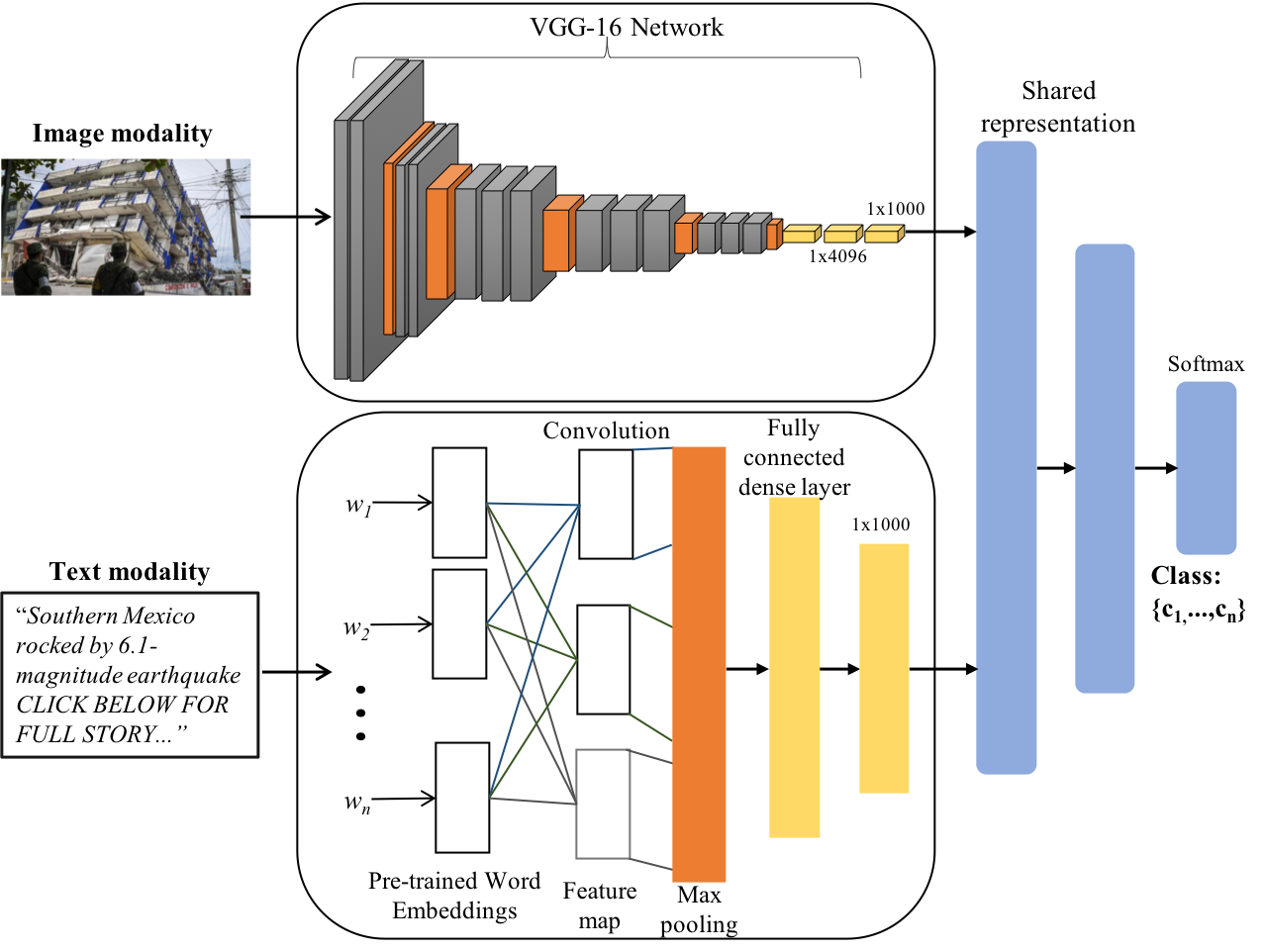}
}
\caption{The multimodal architecture for the classification task using both text and image as input to the system.}
\label{fig:multimodal}
\end{figure*}

In Figure \ref{fig:multimodal}, we present the architecture of the multimodal deep neural network that we use for the experiment. As can be seen in the figure, for the image modality we use the VGG16 network. For the text modality, we use a CNN based architecture. Before forming the shared representations from both modalities we have another hidden layer of size 1,000 from each side. The reason to choose the same size is to have an equal contribution from both modalities. In the current experimental setting, there is no specific reason for choosing the size of 1,000, which can be optimized empirically. After the concatenation of both modalities, we have one hidden layer before the softmax layer.

We use the Adam optimizer with a minibatch size of $32$ for training the model. In order to avoid overfitting, we use early-stopping condition, and as an activation function, we choose ReLU. For this experiment, we do not tune any hyper-parameter (e.g., the size of hidden layers, filter size, dropout rate, etc.). Hence, there is room for further improvement in future studies.

\subsection{Results}
\label{ssec:baseline_results}
In Tables \ref{table:classification_results_info} and \ref{table:classification_results_hum}, we present the performance results achieved for different tasks and modalities. In the unimodal experiments, the image-only models perform better than the text-only models in both informativeness and humanitarian tasks. Specifically, the improvement is more than 2\% on average in the informativeness task whereas  it is more than 6\% on average in the humanitarian task. In the multimodal experiments, we observe additional improvements in performance in both tasks. Specifically, multimodal model performs about 1\% better than the image-only model in all measures for the informativeness task and about 2\% better than the image-only model in all measures for the humanitarian task. These results confirm that multimodal learning approach provides further performance improvement over the unimodal learning approach.

Overall performance of the humanitarian classification models is lower than the informativeness classification models due to the relatively more complex nature of the former task. However, it is important to note that the results presented in this study are obtained using basic network architectures and should be considered as a baseline study. 
%%Along with this study and the data sets we plan to release the code for replicating the baseline results and for future improvements.

\begin{table}
\centering
\caption{Results for the informativeness classification task.}
\label{table:classification_results_info}
% \scalebox{0.8}{
% \resizebox{\columnwidth}{!}{%
\begin{tabular}{llcccc}
\toprule
\multicolumn{1}{l}{\textbf{Training mode}} & \multicolumn{1}{l}{\textbf{Modality}} & \multicolumn{1}{c}{\textbf{Accuracy}} & \multicolumn{1}{c}{\textbf{Precision}} & \multicolumn{1}{c}{\textbf{Recall}} & \multicolumn{1}{c}{\textbf{F1-score}} \\ \midrule
\multirow{2}{*}{\textbf{Unimodal}} & Text & 80.8 & 81.0 & 81.0 & 80.9 \\
 & Image & 83.3 & 83.1 & 83.3 & 83.2 \\ \midrule
\multirow{1}{*}{\textbf{Multimodal}} 
 & Text + Image & 84.4 & 84.1 & 84.0 & 84.2 \\ \bottomrule
\end{tabular}
% }
\end{table}

\begin{table}
\centering
\caption{Results for the humanitarian classification task.}
\label{table:classification_results_hum}
% \scalebox{0.8}{
% \resizebox{\columnwidth}{!}{%
\begin{tabular}{llcccc}
\toprule
\multicolumn{1}{c}{\textbf{Training mode}} & \multicolumn{1}{l}{\textbf{Modality}} & \multicolumn{1}{c}{\textbf{Accuracy}} & \multicolumn{1}{c}{\textbf{Precision}} & \multicolumn{1}{c}{\textbf{Recall}} & \multicolumn{1}{c}{\textbf{F1-score}} \\ \midrule
\multirow{2}{*}{\textbf{Unimodal}} & Text & 70.4 & 70.0 & 70.0 & 67.7 \\
 & Image & 76.8 & 76.4 & 76.8 & 76.3 \\ \midrule
\multirow{1}{*}{\textbf{Multimodal}}
 & Text + Image & 78.4 & 78.5 & 78.0 & 78.3 \\ \bottomrule
\end{tabular}
% }
\end{table}

\section{Discussion}
\label{sec:discussion}

Deep learning models are data hungry. Given our initial condition to consider only the tweet text and image pairs that have the same labels for a given task, we are left with a limited subset of the original CrisisMMD dataset.
% Given our limited labeled dataset, the proposed data pairing approach shows a good alternative. However, pairing requires intelligent sampling. In this work, we used random sampling, however, a better approach can be used.
The proposed multimodal joint learning showed comparable performance over unimodal models for both text and image modalities.
% However, when one of the modalities is missing in the shared representation architecture, it hinders the model's performance. This we consider as future work. Moreover, going beyond sampling of 10k instances for each class to e.g., 100k/class is a potential area that we will explore further, however, doing so would require more computing resources. 
Furthermore, while designing a model for multiple modalities an important challenge is to find concatenation strategies that better capture important information from both modalities. Towards this direction, we design the model by concatenating hidden layers into another layer to form a joint shared representation. 
% In terms of sampling strategies for the training dataset, random sampling is a promising approach as a preliminary study, however, other approaches that we aim to address are finding instances that can capture within-class variations and are distinctive across classes.

\begin{table}
\small
\centering
\caption{Confusion matrices resulted for the informativeness task: \underline{Inf}ormative, \underline{Not-inf}ormative.}
\label{table:conf_inf}
\resizebox{\columnwidth}{!}{%
\subfloat[Text-only][Text-only]{
\begin{tabular}{lc|cc}
& &\multicolumn{2}{c}{\textit{Predicted}} \\%
%\cline{2-3}
& & Inf & Not-inf \\
\hline
\multirowcell{2}{\textit{Human}}&
Inf & 875 & 155 \\
& Not-inf & 139 & 365 \\
\end{tabular}}
\qquad
\subfloat[Image-only][Image-only]{
\begin{tabular}{lc|cc}
& &\multicolumn{2}{c}{\textit{Predicted}} \\%
& & Inf & Not-inf \\
\hline
\multirowcell{2}{\textit{Human}}&
Inf & 916 & 114 \\
& Not-inf & 145 & 359 \\
\end{tabular}}
\qquad
\subfloat[Text + Image][Text + Image]{
\begin{tabular}{lc|cc}
& &\multicolumn{2}{c}{\textit{Predicted}} \\%
& & Inf & Not-inf \\
\hline
\multirowcell{2}{\textit{Human}}&
Inf & 929 & 101 \\
& Not-inf & 139 & 365 \\
\end{tabular}}%
}
\end{table}

\begin{table}
\centering
\caption{Confusion matrices resulted for the humanitarian task: \underline{A}ffected individuals, \underline{I}nfrastructure and utility damage, \underline{N}ot-humanitarian, \underline{O}ther relevant information, \underline{R}escue, volunteering or donation effort.}
\label{table:conf_hum}
\resizebox{\columnwidth}{!}{%
\subfloat[Text-only][Text-only]{
\begin{tabular}{cc|ccccc}
& &\multicolumn{5}{c}{\textit{Predicted}} \\%
& & A & I & N & O & R \\
\hline
\multirowcell{5}{\textit{Human}}
& A & 0 & 0 & 5 & 1 & 3 \\
&I & 0 & 17 & 41 & 12 & 11 \\
&N & 0 & 1 & 458 & 20 & 25 \\
&O & 0 & 6 & 105 & 112 & 12 \\
&R & 0 & 2 & 37 & 2 & 85 \\
\end{tabular}}
\qquad
\subfloat[Image-only][Image-only]{
\begin{tabular}{cc|ccccc}
& &\multicolumn{5}{c}{\textit{Predicted}} \\%
& & A & I & N & O & R \\
\hline
\multirowcell{5}{\textit{Human}}
&A & 1 & 0 & 4 & 0 & 4 \\
&I & 1 & 56 & 13 & 6 & 5 \\
&N & 0 & 13 & 437 & 22 & 32 \\
&O & 0 & 5 & 50 & 178 & 2 \\
&R & 0 & 5 & 43 & 5 & 73 \\
\end{tabular}}
\qquad
\subfloat[Text + Image][Text + Image]{
\begin{tabular}{cc|ccccc}
& &\multicolumn{5}{c}{\textit{Predicted}} \\%
& & A & I & N & O & R \\
\hline
\multirowcell{5}{\textit{Human}}
&A & 1 & 0 & 3 & 0 & 5 \\
&I & 1 & 61 & 10 & 4 & 5 \\
&N & 0 & 17 & 426 & 26 & 35 \\
&O & 0 & 3 & 49 & 180 & 3 \\
&R & 0 & 9 & 33 & 3 & 81 \\
\end{tabular}}%
}
\end{table}

We further analyzed the performance of the three models (i.e., text-only, image-only, and text + image) by examining their confusion matrices. Table~\ref{table:conf_inf} shows three confusion matrix for the three models for the first task (i.e., informative vs. not-informative). From the emergency managers' point of view, it is important that the machine does not miss any useful and relevant message/tweet. The three confusion matrices (a, b, \& c) reveal that our text-only and image-only models missed 155 and 114 instances, respectively, whereas multimodal model missed only 101 instances. These are the instances where machine says ``not informative'', but the ground-truth labels (i.e., human annotators) say ``informative''  (a.k.a. false negatives). The image-only model made significant improvements over the text-only model, however, when text and image modalities are combined in the multimodal case, the error rate dropped significantly (i.e., from 155 to 101).

Another important aspect is related to information overload on emergency managers during a disaster situation. Specifically, it happens when the machine says a message is informative, but according to ground-truth labels it is not (a.k.a. false positives). The confusion matrices in Table~\ref{table:conf_inf} show these mistakes made by the three models as 139 by the text-only, 145 by image-only, and 139 in the multimodal case. We do not observe any improvements from the multimodal approach as observed in the false negative case. 

Table~\ref{table:conf_hum} shows confusion matrices from the three models for the humanitarian categorization tasks. One prominent and important column to observe is ``N'', which corresponds to the ``not-humanitarian'' category, and shows all instances where the model prediction is ``not-humanitarian''. 
\textcolor{black}{In particular, if we look at the number of instances where actual label is ``infrastructure and utility damage'' (denoted as ``I'') but the model prediction is ``not-humanitarian'' (i.e., the value of the cell at the intersection of row ``I'' and column ``N''), we see that the text-only model has 41 false negative instances in Table~\ref{table:conf_hum}(a) whereas the image-only and multimodal models have 13 and 10 instances in Tables~\ref{table:conf_hum}(b) and~\ref{table:conf_hum}(c), respectively. This indicates that the image modality helps models better understand the ``infrastructure and utility damage'' category, and hence, significantly reduce the errors in the predictions. A similar phenomenon can be observed in favor of the text modality for the case where the actual label is ``rescue, volunteering or donation effort'' (denoted as ``R'') whereas the predicted label is ``not-humanitarian'' (i.e., the value of the cell at the intersection of row ``R'' and column ``N''). Specifically, the image-only model has 43 false negative instances in Table~\ref{table:conf_hum}(b) while the text-only and multimodal models have 37 and 33 instances in Tables~\ref{table:conf_hum}(a) and~\ref{table:conf_hum}(c), respectively. In general, we see that the number of such errors are minimized by the third model which uses both text and image modalities together.}
% The cell that intersects with ``I'' (i.e., infrastructure and utility damage) and ``N'' shows 41 false negative instances in matrix `a', 13 in matrix `b', and 10 in matrix `c', which is a clear indication that the image modality helps models understand the category of infrastructure damage. Similar, but a little less improvement can also be observed where row `R' and column `N' intersects for all three matrices. The least number of errors are made by the 3rd model, which uses text and image modalities together. 
However, there are still some cases where significant improvements can be achieved. For instance, the ``other relevant information'' category (denoted as ``O'') seems to create confusion for all the models, which needs to be investigated in a more detailed study.
% For instance, the `O' column that represents the ``Other relevant information'' category creates confusion for all the models.

\begin{figure*}[htbp!]
	\renewcommand{\arraystretch}{0.6} % this reduces the vertical spacing between rows
	\linespread{0.5}\selectfont\centering
	\resizebox{0.99\linewidth}{!}{%
		\begin{tabular}{p{0.35\textwidth} p{0.35\textwidth} p{0.35\textwidth}}
			\includegraphics[width=0.35\textwidth]{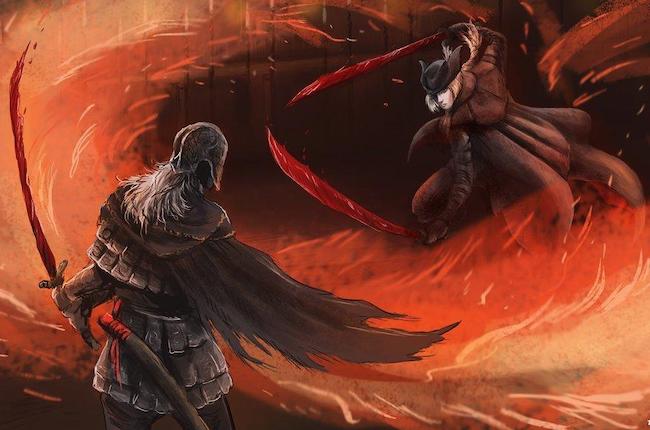}
			&
			\includegraphics[width=0.35\textwidth]{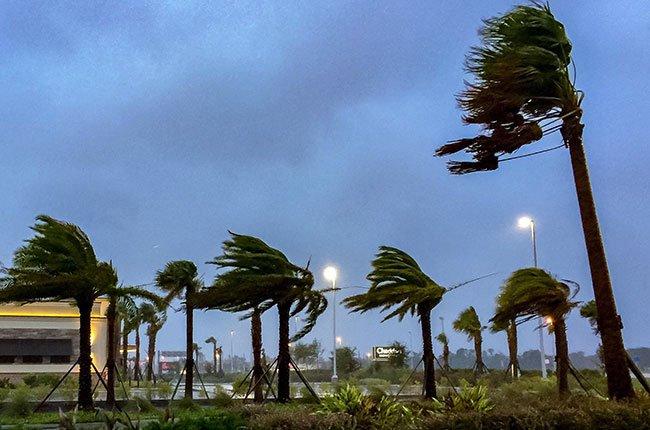}
			&
			\includegraphics[width=0.35\textwidth]{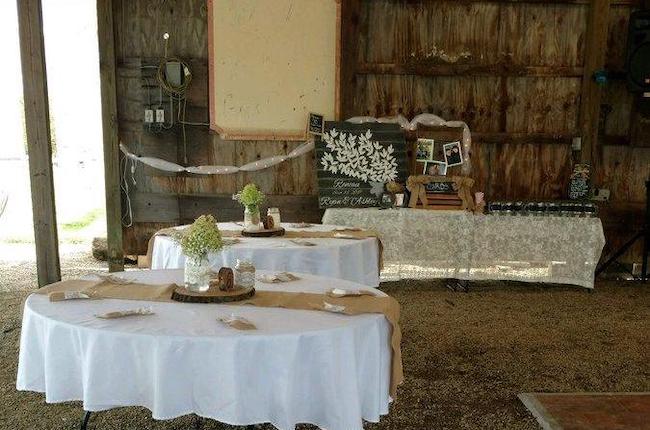}
			\\
% 			{\scriptsize \textbf{(a)} {@}SockyNoob Hurricane Lady \#Maria It'll rain burning blood. I hope Puerto Rico knows how to do Visceral Attacks. https://t.co/HYzgKlwkmz}
			{\scriptsize \textbf{(a)} <USER> Hurricane Lady \#Maria It'll rain burning blood. I hope Puerto Rico knows how to do Visceral Attacks.}
			&
% 			{\scriptsize \textbf{(b)} Hurricane Irma: Rapid response team `rescues' fine wines - https://t.co/pUEeOixSdc \#finewine \#HurricaneIrma https://t.co/JMvGSZJCQL}
			{\scriptsize \textbf{(b)} Hurricane Irma: Rapid response team `rescues' fine wines - https://t.co/pUEeOixSdc \#finewine \#HurricaneIrma}
			&
% 			{\scriptsize \textbf{(c)} RT TracyCarloss: Hurricane Irma nearly ruins a wedding day here in northeast Ohio! Social meeting comes to the res... https://t.co/AdIK9ZZaJs}
			{\scriptsize \textbf{(c)} RT <USER>: Hurricane Irma nearly ruins a wedding day here in northeast Ohio! Social meeting comes to the rescue}
			\\
			{\scriptsize \textbf{Unimodal:} informative (\xmark)}
			&
			{\scriptsize \textbf{Unimodal:} not-informative (\xmark)}
			&
			{\scriptsize \textbf{Unimodal:} informative (\xmark)}
			\\
			{\scriptsize \textbf{Multimodal:} not-informative (\cmark)}
			&
			{\scriptsize \textbf{Multimodal:} informative (\cmark)}
			&
			{\scriptsize \textbf{Multimodal:} not-informative (\cmark)}
			\\
			\includegraphics[width=0.35\textwidth]{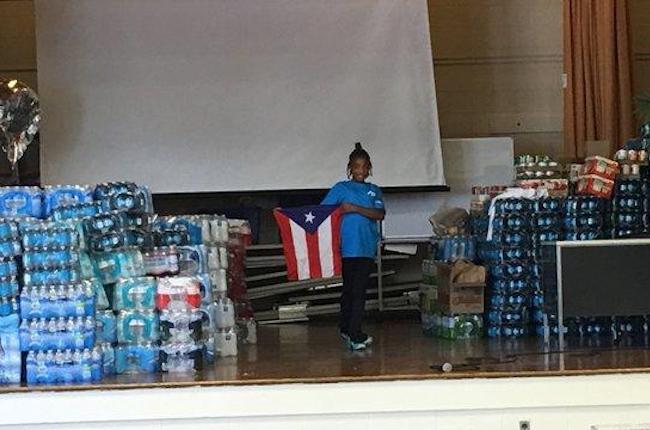}
			&
			\includegraphics[width=0.35\textwidth]{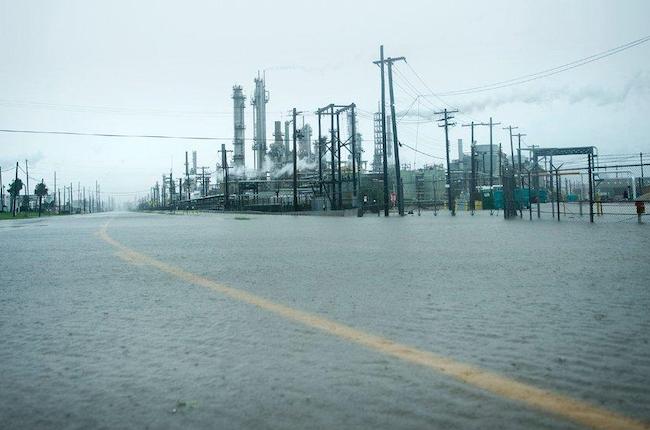}
			&
			\includegraphics[width=0.35\textwidth]{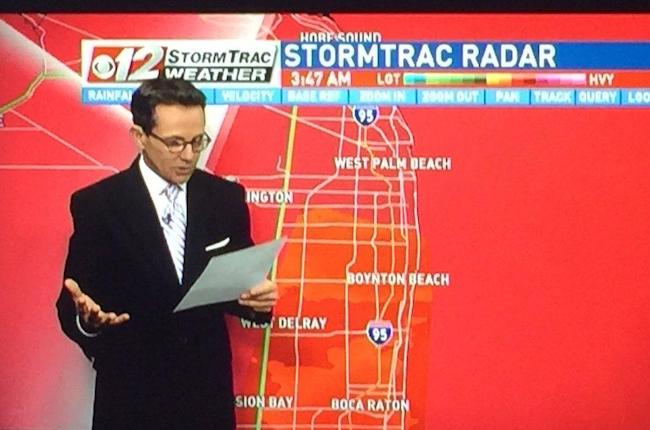}
			\\
% 			{\scriptsize \textbf{(d)} 6th grade Maryland student collects 3,000 cases of drinking water for Puerto Rico https://t.co/x57AeLHeaC https://t.co/ljsqlBiCBh}
			{\scriptsize \textbf{(d)} 6th grade Maryland student collects 3,000 cases of drinking water for Puerto Rico https://t.co/x57AeLHeaC}
			&
% 			{\scriptsize \textbf{(e)} Hurricane Harvey's impact on the US oil industry https://t.co/zxVWR3u0fU https://t.co/lNH5Z1uhlZ}
			{\scriptsize \textbf{(e)} Hurricane Harvey's impact on the US oil industry https://t.co/zxVWR3u0fU}
			&
% 			{\scriptsize \textbf{(f)} \#Breaking Tornado warning for Lantana Rd south to Boca Raton. \#BeSafe {@}CBS12 https://t.co/kg5dveoRxt}
			{\scriptsize \textbf{(f)} \#Breaking Tornado warning for Lantana Rd south to Boca Raton. \#BeSafe <USER>}
			\\
			{\scriptsize \textbf{Text-only:} not-humanitarian (\xmark)}
			&
			{\scriptsize \textbf{Text-only:} other relevant information (\xmark)}
			&
			{\scriptsize \textbf{Text-only:} not-humanitarian (\xmark)}
			\\
			{\scriptsize \textbf{Image-only:} not-humanitarian (\xmark)}
			&
			{\scriptsize \textbf{Image-only:} not-humanitarian (\xmark)}
			&
			{\scriptsize \textbf{Image-only:} not-humanitarian (\xmark)}
			\\
			{\scriptsize \textbf{Multimodal:} rescue, volunteering or donation effort (\cmark)}
			&
			{\scriptsize \textbf{Multimodal:} infrastructure and utility damage (\cmark)}
			&
			{\scriptsize \textbf{Multimodal:} other relevant information (\cmark)}
			\\
		\end{tabular}
	}
	\caption{Example tweet text and image pairs where joint modeling of the input modalities yield better predictions. The symbol (\xmark) indicates an incorrect prediction and the symbol (\cmark) indicates a correct prediction.}
	\label{fig:example_results}
	%\vspace{-.2in}
\end{figure*}

\textcolor{black}{Figure~\ref{fig:example_results} shows example image and text pairs that illustrate how joint modeling of image and text modalities can yield better predictions, and hence, lead to performance improvements over unimodal models. For instance in (a), we reckon that the image-only model thinks the image is informative because it shows \emph{fire-like} patterns whereas the text-only model thinks the text is informative because it mentions \emph{rain burning blood}. However, when these two modalities are evaluated together, they do \emph{not} really provide any consistent evidence for the model to label this image-text pair as informative any more. Similarly, in (d), evaluating image alone or text alone does not result in correct predictions whereas joint evaluation of image and text yields the correct label, i.e., ``rescue, volunteering or donation effort''. Furthermore, we observe another interesting case in (e): text-only model thinks there is potentially useful information for humanitarian purposes by predicting ``other relevant information'' whereas the image-only model thinks there is nothing related to humanitarian purposes by predicting ``not-humanitarian''. However, the multimodal model effectively fuses the information coming from both modalities to make the correct prediction, i.e., ``infrastructure and utility damage''. We believe these examples provide further insights about the success of the proposed multimodal approach for modeling crisis-related social media data.}

\subsection{Challenges and Future Work}

\textcolor{black}{In contrast to other popular multimodal tasks such as image captioning or visual question answering where there is strong alignment or coupling between text and image modalities, social media data are not warranted to have such strong alignment or coupling between co-occuring text and image pairs. In some cases, each modality conveys different type of information, which may even be contradicting the other modality. Therefore, it is important \emph{not} to assume the existence of strong correspondences between social media text and images. To this date, this is a relatively less explored phenomenon that needs more attention from the research community since all of the existing multimodal classification approaches assume that there always exists a common label for data coming from different modalities.} As such, a challenging future direction is to design a multimodal learning algorithm that can be also trained on heterogeneous input, i.e., tweet text and image pairs with disagreeing labels, in which case CrisisMMD can be used for model training in its full form.
\section{Conclusion}
\label{sec:conclusion}

Important informative signals gathered from different data modalities on social media can be highly useful for humanitarian organizations for disaster response. Although images shared on social media contain useful information, past studies have largely focused on text analysis, let alone combining both modalities to get better performance. In this work, we proposed to learn a joint representation using both text and image modalities of social media data. Specifically, we use state-of-the-art deep learning architectures to learn high-level feature representations from text and images to perform two classification tasks. Several experiments performed on real-world disaster-related datasets reveal the usefulness of the proposed approach. 
\textcolor{black}{In summary, our study has two main contributions: (i) It provides baseline results, all in one place, using unimodal and multimodal approaches for both informativeness and humanitarian tasks on the CrisisMMD dataset, and (ii) it shows that a feature-fusion-based multimodal deep neural network can outperform the unimodal models on the challenging CrisisMMD dataset for both tasks, which underlines the importance of multimodal analysis of the crisis-related social media data.} 
Despite the fact that our multimodal classifiers achieve better performance than the unimodal classifiers, we remark that there is still big room for improvement, which we leave for future work.

\printbibliography

@inproceedings{deng2009imagenet,
  title={Imagenet: A large-scale hierarchical image database},
  author={Deng, Jia and Dong, Wei and Socher, Richard and Li, Li-Jia and Li, Kai and Fei-Fei, Li},
  booktitle={2009 IEEE conference on computer vision and pattern recognition},
  pages={248--255},
  year={2009},
  organization={Ieee}
}

@inproceedings{gautam2019multimodal,
  title={Multimodal Analysis of Disaster Tweets},
  author={Gautam, Akash Kumar and Misra, Luv and Kumar, Ajit and Misra, Kush and Aggarwal, Shashwat and Shah, Rajiv Ratn},
  booktitle={2019 IEEE Fifth International Conference on Multimedia Big Data (BigMM)},
  pages={94--103},
  year={2019},
  organization={IEEE}
}

@book{kuncheva2004combining,
  title={Combining pattern classifiers: methods and algorithms},
  author={Kuncheva, Ludmila I},
  year={2004},
  publisher={John Wiley \& Sons}
}

@article{alam2019SocialMedia,
author = {Firoj Alam and Ferda Ofli and Muhammad Imran},
title = {Processing Social Media Images by Combining Human and Machine Computing during Crises},
journal = {International Journal of Human–Computer Interaction},
volume = {34},
number = {4},
pages = {311-327},
year  = {2018},
publisher = {Taylor & Francis},
doi = {10.1080/10447318.2018.1427831},
}

@inproceedings{ngiam2011multimodal,
  title={Multimodal deep learning},
  author={Ngiam, Jiquan and Khosla, Aditya and Kim, Mingyu and Nam, Juhan and Lee, Honglak and Ng, Andrew Y},
  booktitle={Proceedings of the 28th international conference on machine learning (ICML-11)},
  pages={689--696},
  year={2011}
}

@INPROCEEDINGS{alam2014, 
author={F. {Alam} and G. {Riccardi}}, 
booktitle={2014 IEEE International Conference on Acoustics, Speech and Signal Processing (ICASSP)}, 
title={Fusion of acoustic, linguistic and psycholinguistic features for Speaker Personality Traits recognition}, 
year={2014}, 
volume={}, 
number={}, 
pages={955-959}, 
doi={10.1109/ICASSP.2014.6853738}, 
ISSN={1520-6149}, 
month={5},}

@article{Chowdhury2019,
author = {Chowdhury, Shammur Absar and Stepanov, Evgeny A. and Danieli, Morena and Riccardi, Giuseppe},
doi = {10.1016/j.csl.2018.12.001},
issn = {10958363},
journal = {Computer Speech and Language},
pages = {145--167},
publisher = {Elsevier Ltd},
title = {{Automatic classification of speech overlaps: Feature representation and algorithms}},
url = {https://doi.org/10.1016/j.csl.2018.12.001},
volume = {55},
year = {2019}
}

@article{Nagrani2018,
author = {Nagrani, Arsha and Albanie, Samuel and Zisserman, Andrew},
doi = {10.1109/CVPR.2018.00879},
isbn = {9781538664209},
issn = {10636919},
journal = {Proceedings of the IEEE Computer Society Conference on Computer Vision and Pattern Recognition},
pages = {8427--8436},
title = {{Seeing Voices and Hearing Faces: Cross-Modal Biometric Matching}},
year = {2018}
}

@inproceedings{Pereira2016,
author = {Pereira, Mois{\'{e}}s H. R. and P{\'{a}}dua, Fl{\'{a}}vio L. C. and Pereira, Adriano C. M. and Benevenuto, Fabr{\'{i}}cio and Dalip, Daniel H.},
booktitle = {Tenth International AAAI Conference on Web and Social Media (ICWSM)},
pages = {659--662},
title = {Fusing Audio, Textual and Visual Features for Sentiment Analysis of News Videos},
year = {2016}
}

@article{Poria2016,
author = {Poria, Soujanya and Cambria, Erik and Howard, Newton and Huang, Guang Bin and Hussain, Amir},
doi = {10.1016/j.neucom.2015.01.095},
issn = {18728286},
journal = {Neurocomputing},
pages = {50--59},
publisher = {Elsevier},
title = {{Fusing audio, visual and textual clues for sentiment analysis from multimodal content}},
url = {http://dx.doi.org/10.1016/j.neucom.2015.01.095},
volume = {174},
year = {2016}
}

@article{Mouzannar2018,
author = {Mouzannar, Hussein and Rizk, Yara and Awad, Mariette},
journal = {15th International Conference on Information Systems for Crisis Response and Management (ISCRAM 2018)},
number = {May},
pages = {529--543},
title = {{Damage Identification in Social Media Posts using Multimodal Deep Learning}},
year = {2018}
}

@inproceedings{nguyen2017robust,
	Author = {Nguyen, Dat Tien and Al-Mannai, Kamla and Joty, Shafiq R and Sajjad, Hassan and Imran, Muhammad and Mitra, Prasenjit},
	Booktitle = {ICWSM},
	Pages = {632--635},
	Title = {Robust Classification of Crisis-Related Data on Social Networks Using Convolutional Neural Networks.},
	Year = {2017}}

@inbook{CarlosCastillo2016,
	Address = {Istanbul},
	Author = {Castillo, Carlos and Imran, Muhammad and Meier, Patrick and Lucas, Ji Kim and Srivastava, Jaideep and Leson, Heather and Ofli, Ferda and Mitra, Prasenjit},
	Editor = {OCHA and partners},
	Pages = {93-95},
	Publisher = {Tudor Rose, World Humanitarian Summit},
	Title = {Together We Stand---Supporting Decision in Crisis Response: Artificial Intelligence for Digital Response and MicroMappers},
	Year = {2016}}

@inproceedings{szegedy2015going,
	Author = {Szegedy, Christian and Liu, Wei and Jia, Yangqing and Sermanet, Pierre and Reed, Scott and Anguelov, Dragomir and Erhan, Dumitru and Vanhoucke, Vincent and Rabinovich, Andrew},
	Booktitle = {Proceedings of the IEEE conference on computer vision and pattern recognition},
	Pages = {1--9},
	Title = {Going deeper with convolutions},
	Year = {2015}}

@inproceedings{krizhevsky2012imagenet,
	Author = {Krizhevsky, Alex and Sutskever, Ilya and Hinton, Geoffrey E},
	Booktitle = {Advances in neural information processing systems},
	Pages = {1097--1105},
	Title = {{ImageNet} classification with deep convolutional neural networks},
	Year = {2012}}

@article{imran2015processing,
	Author = {Imran, Muhammad and Castillo, Carlos and Diaz, Fernando and Vieweg, Sarah},
	Journal = {ACM Computing Surveys},
	Number = {4},
	Pages = {67},
	Publisher = {ACM},
	Title = {Processing social media messages in mass emergency: A survey},
	Volume = {47},
	Year = {2015}}

@inproceedings{petersinvestigating,
	Author = {Peters, Robin and Joao, Porto de Albuqerque},
	Booktitle = {International Conference on Information Systems for Crisis Response and Management},
	Title = {Investigating images as indicators for relevant social media messages in disaster management},
	Year = {2015}}

@article{simonyan2014very,
	Author = {Simonyan, Karen and Zisserman, Andrew},
	Journal = {arXiv preprint arXiv:1409.1556},
	Title = {Very deep convolutional networks for large-scale image recognition},
	Year = {2014}}

@inproceedings{yosinski2014transferable,
	Acmid = {2969197},
	Author = {Yosinski, Jason and Clune, Jeff and Bengio, Yoshua and Lipson, Hod},
	Booktitle = {Advances in Neural Information Processing Systems},
	Numpages = {9},
	Pages = {3320--3328},
	Title = {How Transferable Are Features in Deep Neural Networks?},
	Year = {2014}}

@inproceedings{ozbulak2016transferable,
	Author = {G. Ozbulak and Y. Aytar and H. K. Ekenel},
	Booktitle = {International Conference of the Biometrics Special Interest Group},
	Doi = {10.1109/BIOSIG.2016.7736925},
	Month = {9},
	Pages = {1-6},
	Title = {How Transferable Are CNN-Based Features for Age and Gender Classification?},
	Year = {2016}}

@inproceedings{nguyen2017automatic,
	Author = {Nguyen, Dat Tien and Alam, Firoj and Ofli, Ferda and Imran, Muhammad},
	Booktitle = {International Conference on Information Systems for Crisis Response and Management ({ISCRAM})},
	Month = {5},
	Title = {Automatic Image Filtering on Social Networks Using Deep Learning and Perceptual Hashing During Crises},
	Year = {2017}}

@inproceedings{alam17demo,
	Author = {Alam, Firoj and Imran, Muhammad and Ofli, Ferda},
	Booktitle = {International Conference on Advances in Social Networks Analysis and Mining ({ASONAM})},
	Month = {8},
	Pages = {1--4},
	Title = {Image4Act: Online Social Media Image Processing for Disaster Response.},
	Year = {2017}}

@inproceedings{daly2016mining,
	Author = {Daly, Shannon and Thom, J},
	Booktitle = {International Conference on Information Systems for Crisis Response and Management},
	Pages = {1--14},
	Title = {Mining and Classifying Image Posts on Social Media to Analyse Fires},
	Year = {2016}}

@inproceedings{Alam2018GraphBS,
	Author = {Firoj Alam and Shafiq R. Joty and Muhammad Imran},
	Booktitle = {Proc. of the 12th ICWSM, 2018},
	Publisher = {AAAI press},
	Title = {Graph Based Semi-supervised Learning with Convolution Neural Networks to Classify Crisis Related Tweets},
	Year = {2018}}

@article{srivastava2014dropout,
	Author = {Srivastava, Nitish and Hinton, Geoffrey E and Krizhevsky, Alex and Sutskever, Ilya and Salakhutdinov, Ruslan},
	Journal = {Journal of MLR},
	Number = {1},
	Pages = {1929--1958},
	Title = {Dropout: a simple way to prevent neural networks from overfitting.},
	Volume = {15},
	Year = {2014}}

@article{ioffe2015batch,
	Author = {Ioffe, Sergey and Szegedy, Christian},
	Journal = {arXiv preprint arXiv:1502.03167},
	Title = {Batch normalization: Accelerating deep network training by reducing internal covariate shift},
	Year = {2015}}

@article{zeiler2012adadelta,
	Author = {Zeiler, Matthew D},
	Journal = {arXiv preprint arXiv:1212.5701},
	Title = {ADADELTA: an adaptive learning rate method},
	Year = {2012}}

@inproceedings{alam2018crisismmd,
	Author = {Firoj Alam and Ferda Ofli and Muhammad Imran},
	Booktitle = {Proc. of the 12th ICWSM, 2018},
	Day = {1},
	Language = {English},
	Month = {1},
	Pages = {465--473},
	Publisher = {AAAI press},
	Title = {{CrisisMMD}: Multimodal twitter datasets from natural disasters},
	Year = {2018}}

@inproceedings{bica2017visual,
	Author = {Bica, Melissa and Palen, Leysia and Bopp, Chris},
	Booktitle = {Proc. of the CSCW},
	Pages = {1262--1276},
	Title = {Visual Representations of Disaster.},
	Year = {2017}}

@inproceedings{rosenthal2017semeval,
	Author = {Rosenthal, Sara and Farra, Noura and Nakov, Preslav},
	Booktitle = {Proc. of the 11th SemEval, 2017)},
	Pages = {502--518},
	Title = {SemEval-2017 task 4: Sentiment analysis in Twitter},
	Year = {2017}}

@article{mikolov2013efficient,
	Author = {Mikolov, Tomas and Chen, Kai and Corrado, Greg and Dean, Jeffrey},
	Journal = {arXiv preprint arXiv:1301.3781},
	Title = {Efficient estimation of word representations in vector space},
	Year = {2013}}

@inproceedings{chen2013understanding,
	Author = {Chen, Tao and Lu, Dongyuan and Kan, Min-Yen and Cui, Peng},
	Booktitle = {ACM International Conference on Multimedia},
	Pages = {781--784},
	Title = {Understanding and classifying image tweets},
	Year = {2013}}

@inproceedings{nguyen17damage,
	Author = {Nguyen, Dat Tien and Ofli, Ferda and Imran, Muhammad and Mitra, Prasenjit},
	Booktitle = {International Conference on Advances in Social Networks Analysis and Mining (ASONAM)},
	Month = {8},
	Pages = {1--8},
	Title = {Damage Assessment from Social Media Imagery Data During Disasters},
	Year = {2017}}

\end{document}